\documentclass[10pt,twocolumn,letterpaper]{article}

\usepackage{cvpr}      

\usepackage[table]{xcolor}

\usepackage{etoolbox} 
\AtBeginEnvironment{table}{\small} 
\usepackage{booktabs}

\usepackage{tabularx}
\usepackage{booktabs}
\usepackage{array}

\definecolor{cvprblue}{rgb}{0.21,0.49,0.74}
\usepackage[pagebackref,breaklinks,colorlinks,allcolors=cvprblue]{hyperref}

\def\paperID{*****} 
\def\confName{CVPR}
\def\confYear{2026}

\title{AITP: Traffic Accident Responsibility Allocation \\
via Multimodal Large Language Models}

\author{
Zijin Zhou, Songan Zhang\thanks{Corresponding author.}\\
Global Institute of Future Technology, Shanghai Jiao Tong University\\
{\tt\small 787796134@sjtu.edu.cn, songanz@sjtu.edu.cn}\\
{\tt\small \url{https://github.com/zijinzhou2005/AITP}}
}

\begin{document}
\maketitle
\begin{abstract}
Multimodal Large Language Models (MLLMs) have achieved remarkable progress in Traffic Accident Detection (TAD) and Traffic Accident Understanding (TAU). 
However, existing studies mainly focus on describing and interpreting accident videos, leaving room for deeper causal reasoning and integration of legal knowledge. 
Traffic Accident Responsibility Allocation (TARA) is a more challenging task that requires multi-step reasoning grounded in traffic regulations. 
To address this, we introduce \textbf{AITP} (Artificial Intelligence Traffic Police), a multimodal large language model for responsibility reasoning and allocation. 
AITP enhances reasoning via a Multimodal Chain-of-Thought (MCoT) mechanism and integrates legal knowledge through Retrieval-Augmented Generation (RAG). 
We further present \textbf{DecaTARA}, a decathlon-style benchmark unifying ten interrelated traffic accident reasoning tasks with 67,941 annotated videos and 195,821 question–answer pairs. 
Extensive experiments show that AITP achieves state-of-the-art performance across responsibility allocation, TAD, and TAU tasks, establishing a new paradigm for reasoning-driven multimodal traffic analysis.
\end{abstract}
    
\section{Introduction}
\label{sec:intro}
Traffic accidents occur in large numbers every year, posing a significant threat to public safety and traffic management\cite{who2023roadsafety}. Currently, traffic accident responsibility allocation (TARA) still relies mainly on manual judgment, which is both time-consuming and labor-intensive. Developing an automated TARA system can greatly improve the efficiency of responsibility determination and, to some extent, substitute manual efforts in routine cases. However, research specifically focused on TARA remains largely unexplored.

\begin{figure*}[t]
    \centering
    \includegraphics[width=0.9\linewidth]{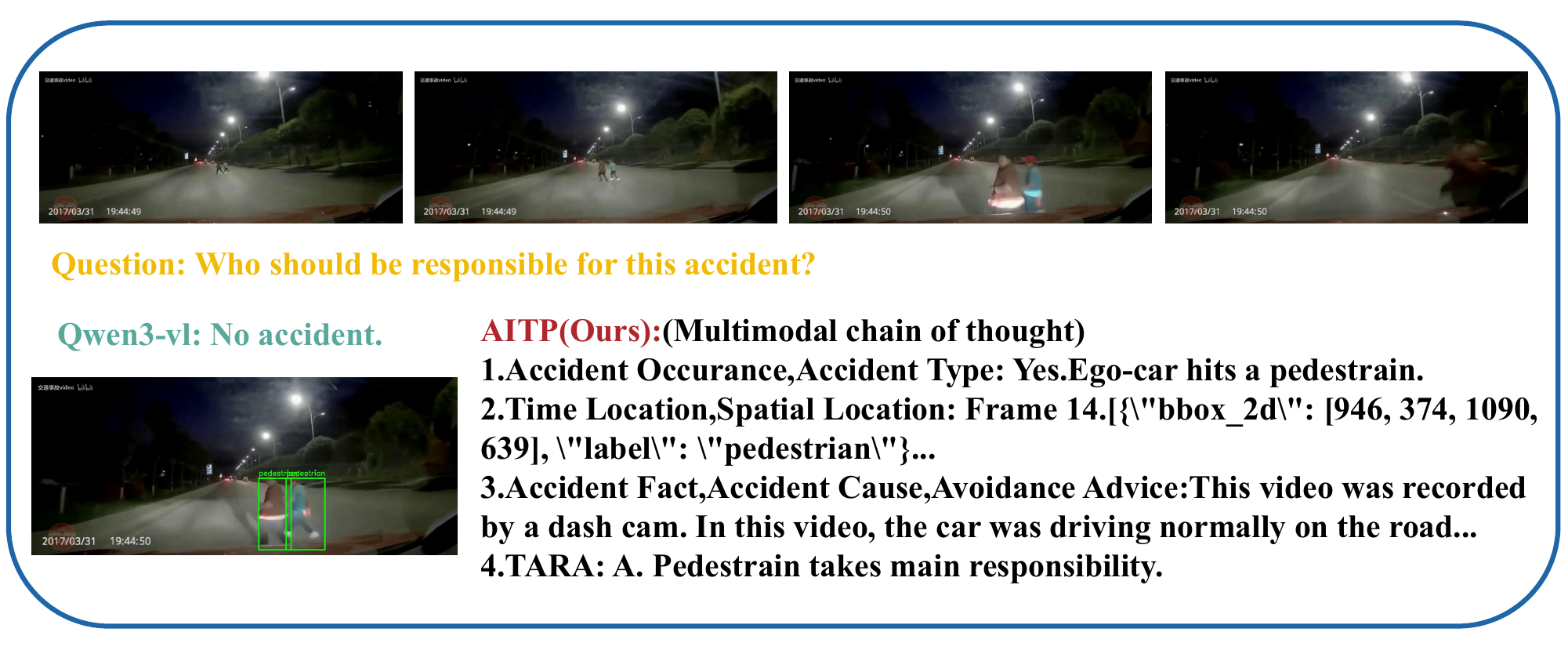}
    \caption{One example to illustrate the limitations of general MLLM in TARA. 
    In the scenario pedestrian was hit while crossing the road. Qwen3-VL fail to detect this issue.}
    \label{figure1}
\end{figure*}

With the rapid development of computer vision and deep learning, earlier studies have achieved automated Traffic Accident Detection (TAD)\cite{fang2022traffic,melegrito2024deep,hajri2022vision,haresh2020towards,roy2020detection}, while recent advances in Multimodal Large Language Models (MLLMs) have shown strong performance in Traffic Accident Understanding (TAU)\cite{Xu_2021_CVPR,Xing_2025_CVPR,zhou2025tauk,sheng2025safeplug,fang2024abductive,moura2025nexar,parikh2025roadsocial,jin2023adapt,yuan2024rag,ma2024dolphins}, particularly in generating textual descriptions of accident scenes. Although existing approaches have achieved remarkable results in both TAD and TAU, they mainly focus on detecting the timing of accidents, identifying the entities involved, and generating textual descriptions of the accident process and its causes. These works have greatly advanced intelligent traffic video understanding but remain at the perceptual and descriptive levels. In contrast, TARA requires a deeper level of reasoning ability—not only to analyze causal relationships and determine responsible parties, but also to comprehend and apply traffic regulations. However, research on enabling multimodal large models to perform such reasoning-driven responsibility determination tasks remains largely unexplored.More importantly,as shown in Fig.~\ref{figure1}, directly applying existing approaches to TARA often leads to hallucination issues, as the models lack structured causal reasoning and external legal knowledge, resulting in unstable or unreliable responsibility judgments.

Meanwhile, despite the growing attention on multimodal models for traffic video analysis in recent years, there is still no dataset specifically designed for TARA. Existing datasets\cite{Xu_2021_CVPR,Xing_2025_CVPR,zhou2025tauk,sheng2025safeplug,fang2024abductive,parikh2025roadsocial,moura2025nexar} mostly focus on accident detection or description tasks, with annotations limited to accident occurrence, categories, and event narratives, lacking systematic labeling of responsibility attribution. Moreover, responsibility determination often involves two participants and complex spatiotemporal interactions, making it challenging to scientifically and objectively annotate responsibility outcomes in traffic accidents. This remains an important yet underexplored problem.

In this work, we present DecaTARA (A Decathlon Dataset for Traffic Accident Responsibility Allocation), the first large-scale multimodal dataset specifically designed for TARA. DecaTARA contains X traffic accident videos and Y question–answer pairs. Building upon previous studies in TAD and TAU, we integrate and unify multiple existing datasets to bridge the gap between accident detection and causal reasoning. To comprehensively enhance the model’s capability in traffic accident understanding, DecaTARA is designed with ten tasks, including seven accident-related tasks, two non-accident tasks, and one evaluation task. The accident-related tasks cover accident occurrence judgment, accident type classification, accident fact description, accident reason explanation, accident temporal and spatial location, and avoidance advice. To prevent overfitting and enhance model generalization, we further construct a non-accident subset containing two tasks that describe and infer vehicle behaviors in normal driving scenarios. Finally, we annotate a responsibility allocation subset that quantitatively evaluates models through responsibility allocation accuracy. To the best of our knowledge, DecaTARA is the first large-scale, multi-task dataset dedicated to traffic accident responsibility allocation.

Secondly, we propose AITP, a multimodal large language model designed for TARA. AITP introduces two core innovations—Multimodal Chain of Thought (MCoT)\cite{wei2022chain,zhang2023multimodal} and Retrieval-Augmented Generation (RAG)\cite{lewis2020retrieval} — which jointly mitigate hallucination issues during responsibility determination. The model performs step-by-step reasoning on accident occurrence, type, time, subject, fact, cause, and avoidance advice, and ultimately completes responsibility allocation and self-verification. All these sub-tasks are fine-tuned on the DecaTARA dataset to enhance the model’s perceptual and reasoning capabilities, while the responsibility allocation task is accomplished through reasoning based on the previous chain. During this TARA process, the RAG module provides the model with relevant traffic regulation knowledge, enabling fact- and law-grounded reasoning for more reliable and interpretable responsibility attribution.

We evaluate our model across multiple tasks in TAD and TAU, and for the first time, introduce responsibility allocation accuracy as a new quantitative metric to more comprehensively assess the model’s understanding and reasoning capability. Experimental results show that AITP achieves state-of-the-art (SOTA) performance on both responsibility allocation accuracy and multi-task benchmarks, significantly outperforming existing baselines. Further ablation studies validate the effectiveness of RAG module, and MCoT reasoning mechanism. 
\noindent\textbf{Our main contributions are summarized as follows:}
\begin{itemize}
    \item We construct \textbf{DecaTARA},the first multi-task dataset for traffic accident responsibility allocation, containing \textit{67,941} videos and \textit{195,821} QA pairs, with ten tasks designed to enhance comprehensive accident understanding.
    
    \item We propose \textbf{AITP}, a multimodal large language model for TARA, which integrates \textbf{MCoT} for stepwise reasoning and \textbf{RAG} for legal knowledge retrieval, effectively reducing hallucination and improving reasoning reliability.
    
    \item Extensive experiments demonstrate that AITP achieves \textbf{SOTA} performance on both responsibility allocation accuracy and multi-task benchmarks.
\end{itemize}
\section{Related Work}
\label{sec:related}

\begin{figure*}[t]
    \centering
    \includegraphics[width=0.9\linewidth]{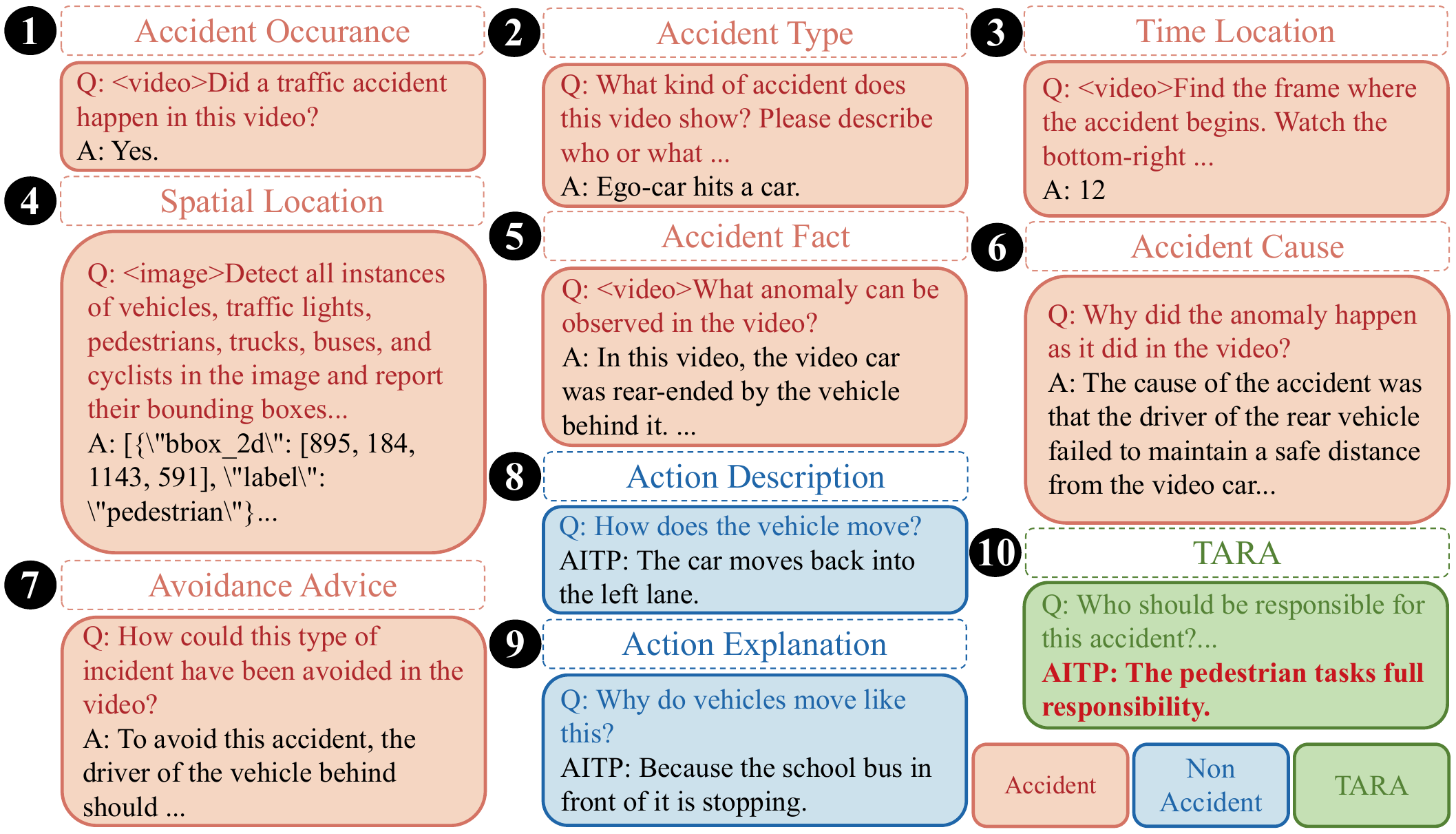}
    \caption{DecaTARA comprises ten tasks: Tasks 1-4 focus on traffic accident detection, Tasks 5-7 are about traffic accident understanding, Tasks 8-9 cover vehicle behavior and cause reasoning in non-accident scenarios, and Task 10 concerns traffic accident responsibility allocation.}
    \label{fig:figure2}
\end{figure*}

\textbf{Traffic Accident Detection and Understanding}.\\
Early traffic accident datasets primarily focused on perception-level tasks such as anomaly detection, type classification, and object localization. Representative dashcam datasets, including A3D\cite{yao2019A3D}, DoTA\cite{yao2022dota}, and CCD\cite{bao2020CCD}, provided temporal and categorical annotations, while DADA\cite{fang2021dada} and DADA-2000\cite{fang2019dada} incorporated driver gaze information and fine-grained semantic segmentation. Surveillance datasets\cite{shah2018cadp,xu2024tad,chai2024tads,lv2021localizing,Xu_2021_CVPR,you2020traffic} offered broader traffic scenes captured by fixed cameras, and synthetic datasets\cite{aliakbarian2018viena,kim2019crash,luo2023simulation,vijay2022detection,wang2024deepaccident,li2025causal,li2025crashagent} achieved larger scale and diversity but still suffered from a notable domain gap compared to real-world scenarios.

In recent years, research attention has gradually shifted toward TAU, emphasizing multimodal reasoning over accident facts, causes, and prevention. SUTD-TrafficQA\cite{Xu_2021_CVPR} pioneered video question-answering for accident understanding, while MM-AU\cite{fang2024abductive} further introduced multi-aspect textual annotations covering factual, causal, and preventive reasoning. TAU-106K\cite{zhou2025tauk} expanded the scale to 106K video–image pairs, enabling comprehensive spatio-temporal grounding and hierarchical understanding. EchoTraffic (AV-TAU)\cite{Xing_2025_CVPR} incorporated audio–visual modalities, leveraging auditory cues such as collisions and braking sounds to enhance multimodal reasoning. SafePLUG\cite{sheng2025safeplug} further introduced pixel-level segmentation and temporal grounding, allowing models to understand accident evolution with fine spatial and temporal granularity.

However, despite the progress of TAU datasets, there remains a lack of benchmarks dedicated to TARA — a more challenging task that requires models to perform causal reasoning and responsibility inference from video evidence. To bridge this gap, we construct DecaTARA, providing a new foundation for developing AI systems capable of automatic accident responsibility determination.

\section{DecaTARA Dataset}
\label{sec:dataset}

We propose DecaTARA, the first large-scale multi-task dataset specifically designed for TARA. DecaTARA contains 67,941 traffic videos and 195,821 question–answer pairs. As shown in Figure~\ref{fig:figure2}, DecaTARA is comprised of ten tasks and consists of three subsets: Accident Dataset, Non-Accident Dataset, and TARA Evaluation Dataset. We perform dataset-specific preprocessing and curation for each source dataset before integration.

To the best of our knowledge, DecaTARA is the largest and most comprehensive dataset for TARA, providing a solid foundation for advancing research in traffic accident understanding and accountability reasoning. Compared with prior evaluation protocols that mostly assess a model’s ability to describe an accident, TARA provides a more systematic framework that quantifies a model’s understanding of the accident. It examines how well the model identifies key agents, their interactions, and responsibility relations.
 
\subsection{Accident Dataset}
To enable the model to achieve a comprehensive and progressive understanding of traffic accident videos, we combine TAD and TAU, designing a seven-task benchmark that covers multiple levels of perception and reasoning.
\subsubsection{Traffic Accident Detection}
For the TAD task, we use the MM-AU~\cite{fang2024abductive} dataset, which contains 11,730 accident videos and 46,920 annotated samples.
To improve category balance and ensure more reliable evaluation, we re-split MM-AU into new training and test subsets according to accident types. Since only frame sequences are publicly available, we reconstruct each sample into a 5-second video by uniformly sampling 40 frames around the accident frame, resulting in videos at 8 FPS. We selected 1,254 videos as the test set.\\
{\bf Task 1: Accident Occurrence.} This subtask requires the model to determine whether a traffic accident occurs in the given video, serving as the foundation for subsequent accident understanding tasks.\\
{\bf Task 2: Accident Type.} This task challenges the model to identify the specific type of accident, such as a vehicle-to-vehicle, vehicle-to-bicycle, or vehicle-to-pedestrian collision.\\
{\bf Task 3: Time Location.} This task requires the model to identify the first frame in which the accident occurs. To facilitate precise temporal grounding, each frame in the input video is annotated with its frame index at the bottom-right corner. The model is explicitly prompted to focus on this region and use the displayed frame index as the temporal reference when determining the accident onset.\\
{\bf Task 4: Spatial Location.} This subtask requires the model to pinpoint the spatial position of the accident subject within the key frame obtained through time localization. The model outputs a bounding box (bbox) indicating the location of the involved object or vehicle.

\begin{table}[h]
\centering
\small
\setlength{\tabcolsep}{6pt}
\begin{tabular}{lccc}
\toprule
\textbf{Source} & \textbf{Videos} & \textbf{Test Videos} & \textbf{Q/A Pairs} \\
\midrule
AV-TAU & 29{,}865 & 3,000 & 89,595 \\
MM-AU  & 11,730 & 1,254 & 46,920 \\
BDDX   & 26,526 & 2,000 & 53,052 \\
TARA Dataset        &    --       &   --      & 6{,}254 \\
\midrule
\textbf{DecaTARA} & \textbf{67,941} & \textbf{6,254} & \textbf{195,821} \\
\bottomrule
\end{tabular}
\caption{Composition of the DecaTARA benchmark. DecaTARA adds responsibility labels to the test splits of existing datasets (no new videos are added).}
\label{tab:tara_dataset}
\end{table}

\subsubsection{Traffic Accident Understanding}
For the TAU task, we employ the AV-TAU~\cite{Xing_2025_CVPR} dataset, which contains 29,865 accident videos and 89,595 corresponding textual annotations. The test set contains 3,000 videos.
Since the original AV-TAU dataset includes descriptions related to audio information, we remove these elements through data cleaning and manual verification to ensure that the dataset focuses purely on visual and textual modalities.\\
{\bf Task 5: Accident Facts.} This task prompts the model to describe the detailed sequence and key events of the accident, demonstrating its capability for fine-grained spatio-temporal understanding.\\
{\bf Task 6: Accident Cause.} This subtask encourages the model to analyze and infer the main cause of the accident based on visual cues and contextual information.\\
{\bf Task 7: Avoidance Advice.} This task prompts the model to provide reasonable and actionable suggestions for avoiding similar accidents in the future.
\subsection{Non-Accident Dataset}
To avoid overfitting and improve the model’s generalization capability, our dataset also incorporates a subset of non-accident videos. These videos help the model learn fundamental traffic knowledge—such as stopping at red lights or moving at green lights—and provide contrastive supervision that enables it to distinguish normal driving behaviors from accident-related anomalies. For BDDX, we segment the original long videos into clips according to the provided temporal annotations. We also correct some unreasonable segment boundaries and refine the annotations through an MLLM-assisted review pipeline followed by manual verification. The non-accident subset contains 26,526 videos and 53,052 video–text pairs.\\
{\bf Task 8: Vehicle Behavior Description.} 
This subtask requires the model to objectively describe the visible behaviors of vehicles in the video, such as acceleration, deceleration, turning, or lane changing, providing essential visual cues for understanding traffic dynamics.\\
{\bf Task 9: Vehicle Behavior Explanation.} 
This subtask requires the model to infer the underlying causes or intentions behind vehicle behaviors—for example, why a vehicle suddenly brakes or deviates from its lane—enabling deeper semantic understanding of traffic scenarios.
\subsection{Traffic Accident Responsibility Allocation Dataset}
To evaluate responsibility reasoning performance, we construct a \textbf{Traffic Accident Responsibility Allocation Dataset}, containing \textbf{6,254} accident videos and an equal number of responsibility annotations. 
The dataset is curated from the test splits of three publicly available sources: 3,000 videos from AV-TAU~\cite{Xing_2025_CVPR}, 1,254 from MM-AU~\cite{fang2024abductive}, and 2,000 from BDDX~\cite{kim2018BDDX}, covering diverse scenarios such as vehicle–pedestrian, vehicle–vehicle, and non-accident.

\vspace{2pt}
\noindent\textbf{Annotation scheme.}
The annotations are categorized into three types:
\begin{itemize}
    \item \textbf{A.} \texttt{[ ]} takes \textbf{main responsibility}.
    \item \textbf{B.} \texttt{[ ]} and \texttt{[ ]} \textbf{share responsibility equally}.
    \item \textbf{C.} \textbf{No accident}.
\end{itemize}

Here, \texttt{[ ]} denotes one of the following entities: pedestrian, motorcycle, bike, vehicle, truck, ego-car, electric bike, bus, taxi, tricycle, van, or tram.  
For instance, “Pedestrian and vehicle share responsibility equally.”  
If multiple entities of the same type are involved, they are distinguished by spatial reference, e.g., “Right vehicle takes main responsibility.”

To maintain quantitative consistency with official traffic fault determination standards, we assign:
\begin{itemize}
    \item subjects with fault ratios above \textbf{70\%} → \textbf{Type A (main responsibility)};
    \item subjects with roughly \textbf{50–60\%} fault each → \textbf{Type B (shared responsibility)}~\cite{awad2018blaming}.
\end{itemize}

Each video is independently annotated by three trained annotators following standardized traffic accident liability criteria. 
Disagreements are resolved through discussion with certified traffic safety specialists to ensure consistency with formal responsibility determination principles.

 \begin{figure*}[t]
    \centering
    \includegraphics[width=0.9\linewidth]{figures/figure3.pdf}
    \caption{Overview of the TARA inference pipeline. The model first performs Multimodal Chain-of-Thought (MCoT) reasoning, sequentially completing Traffic Accident Detection (TAD) and Traffic Accident Understanding (TAU). In time location, the model predicts the accident frame index and extracts the corresponding video frames for evidence grounding. During the final TARA stage, the model retrieves relevant legal articles through the RAG module.}
    \label{fig:aitp_framework}
\end{figure*}
\section{Methodology}
\subsection{Training Strategy}

We adopt a four-stage progressive fine-tuning strategy based on the pretrained Qwen3-VL\cite{bai2025qwen3} to systematically enhance the model’s multimodal understanding and reasoning capabilities in traffic accident scenarios. All four stages are implemented with the Swift\cite{zhao2024swiftascalablelightweightinfrastructure} fine-tuning framework, where each video is uniformly sampled at 8 FPS.\\
{\bf Stage 1 (Traffic Domain Alignment):}
In the first stage, the model is fine-tuned on the non-accident dataset to acquire fundamental knowledge of the traffic domain, such as recognizing vehicles, pedestrians, road signs, and understanding typical behaviors like stopping at red lights, proceeding on green lights, or yielding to pedestrians. This stage aims to align the model with normal traffic patterns and prevent overfitting to accident-related data.
We perform full fine-tuning on the non-accident subset for 1 hour using 8× RTX PRO 6000 GPUs, with a learning rate of 1e-5.\\
{\bf Stage 2 (Traffic Accident Detection – TAD):} In the second stage, the model is fine-tuned on accident-related tasks, focusing on accident detection, type recognition, temporal localization, and spatial localization (corresponding to Tasks 1, 2, 3, and 4 in Fig.~\ref{fig:figure2}). Accident detection and type classification are placed within the same dialogue round, where the first turn focuses on detecting whether an accident occurs and the second turn identifies its type.
To enhance robustness, we mix a small proportion of non-accident videos during TAD training, preventing the model from overfitting to accident-only distributions.
This stage uses full fine-tuning for 3 hours on 8× RTX PRO 6000 GPUs, with a learning rate of 1e-5.\\
{\bf Stage 3 (Traffic Accident Understanding – TAU):}
In the third stage, the model is fine-tuned on higher-level reasoning tasks—accident description, causation analysis, and avoidance suggestion (Tasks 5, 6, and 7 in Fig.~\ref{fig:figure2}).
These subtasks are organized within a single dialogue round to help the model form a coherent reasoning chain from perception to interpretation. We also mix a small proportion of non-accident videos here.
We again apply full fine-tuning for 3 hours on 8× RTX PRO 6000 GPUs, using a learning rate of 1e-5.
Through this stage, the model learns to perform structured reasoning and generate interpretable outputs, serving as the foundation for the subsequent Traffic Accident Responsibility Allocation (TARA) task.\\
{\bf Stage 4 (Traffic Accident Responsibility Allocation – TARA):}
Finally, the model is adapted for responsibility reasoning through LoRA fine-tuning on 1,000 annotated TARA samples.
Because the fully fine-tuned model from Stage 3 tends to drift from instruction adherence, we perform a lightweight LoRA update (rank=8, $\alpha$=32, learning rate=5e-5) to refine its output format and restore controllable response behavior.
This stage is completed within 15 minutes on a single RTX PRO 6000 GPU.

\begin{table*}[t]
\centering
\small
\setlength{\tabcolsep}{4pt}
\caption{Comparison of {\cellcolor{blue!10}\textbf{General MLLMs}} and {\cellcolor{green!10}\textbf{Traffic-specific MLLMs}} across all AITP tasks. 
\textbf{Bold} values indicate the best results.}
\resizebox{\textwidth}{!}{
\begin{tabular}{c c|c c c c|c}
\toprule
\textbf{Task} & \textbf{Metric}
& \multicolumn{4}{c|}{\cellcolor{blue!10}\textbf{General MLLMs}}
& \cellcolor{green!10}\textbf{Traffic-specific} \\
\midrule
& & Gemma-3n-E4B(8B) & Qwen3-VL (8B) & Kimi-VL-A3B(16B) & InternVL3.5(8B) & \textbf{AITP (8B)} \\
\midrule

& Accident & 0.0755 & 0.1895 & 0.0637 & 0.3377 & \textbf{0.7218} \\
\multicolumn{1}{c}{\textbf{TARA}} & Non-Accident & 0.9849 & \textbf{0.9937} & 0.9727 & 0.9795 & 0.8117 \\
& Average & 0.4303 & 0.4956 & 0.4183 & 0.5881 & \textbf{0.7569} \\
\midrule

& BLEU   & 0.0301 & 0.0301 & 0.0219 & 0.0354 & \textbf{0.0395} \\
& ROUGE-L  & 0.1727 & 0.1732 & 0.1433 & 0.1618 & \textbf{0.2027} \\
\multicolumn{1}{c}{\textbf{Description}} & BERTScore   & 0.6349 & 0.6315 & 0.6087 & 0.6029 & \textbf{0.6547} \\
& MoverScore  & 0.5482 & 0.4996 & 0.5320 & 0.3844 & \textbf{0.6628} \\
& GPTEval    & 0.5229 & 0.4052 & 0.5617 & 0.3013 & \textbf{0.6448} \\
\midrule

& BLEU   & 0.0334 & 0.0177 & 0.0257 & 0.0315 & \textbf{0.0382} \\
& ROUGE-L  & 0.1623 & 0.1198 & 0.1406 & 0.1687 & \textbf{0.1800} \\
\multicolumn{1}{c}{\textbf{Causation}} & BERTScore   & 0.6277 & 0.5883 & 0.6001 & 0.6215 & \textbf{0.6360} \\
& MoverScore  & 0.5454 & \textbf{0.6291} & 0.5465 & 0.6182 & 0.6565 \\
& GPTEval    & \textbf{0.5879} & 0.4545 & 0.4553 & 0.2862 & 0.5485 \\
\midrule

& BLEU   & 0.0310 & 0.0394 & 0.0276 & 0.0355 & \textbf{0.0441} \\
& ROUGE-L  & 0.1306 & 0.1676 & 0.1409 & 0.1830 & \textbf{0.1911} \\
\multicolumn{1}{c}{\textbf{Prevention}} & BERTScore   & 0.6337 & 0.6346 & 0.6367 & 0.6180 & \textbf{0.6451} \\
& MoverScore  & 0.6113 & \textbf{0.6340} & 0.6074 & 0.5276 & 0.5673 \\
& GPTEval    & \textbf{0.7560} & 0.5747 & 0.7338 & 0.4609 & 0.6570 \\
\midrule

\multirow{4}{*}{\textbf{Time}} & AP@3 & 0.2442 & 0.1416 & -- & 0.1131 & \textbf{0.5825} \\
& AP@5 & 0.3911 & 0.2368 & -- & 0.2114 & \textbf{0.7992} \\
& AP@7 & 0.5148 & 0.3404 & -- & 0.3330 & \textbf{0.8901} \\
& Average  & 0.3834 & 0.2396 & -- & 0.2192 & \textbf{0.7572} \\
\midrule

\multirow{4}{*}{\textbf{Spatial}} & AP@30 & 0.0496 & 0.0376 & -- & 0.0342 & \textbf{0.6527} \\
& AP@50 & 0.0111 & 0.0068 & -- & 0.0068 & \textbf{0.6291} \\
& AP@70 & 0.0009 & 0.0000 & -- & 0.0009 & \textbf{0.4940} \\
& mIoU  & 0.0645 & 0.0619 & -- & 0.0633 & \textbf{0.6641} \\
\bottomrule
\end{tabular}
}
\label{tab:all_tasks_combined}
\end{table*}

\subsection{Inference Strategy}
\subsubsection{Multimodal chain of thought (MCoT)}

Unlike previous tasks that primarily focus on describing or interpreting video content in textual form, Traffic Accident Responsibility Allocation (TARA) is inherently more complex, as it requires integrating visual and textual evidence, reasoning over inter-object relations, and grounding conclusions in traffic law. To address this challenge, we design a multimodal chain-of-thought (MCoT) framework that conducts a multi-turn dialogue over the same video to progressively accumulate verifiable evidence.

As illustrated in Figure~\ref{fig:aitp_framework}, the dialogue proceeds as follows:

\textbf{(1) Traffic Accident Detection (TAD)}: The model first determines whether an accident occurs. If not, the dialogue terminates. If yes, the next turn classifies the accident type. Then it comes to time location. We overlay the current frame timestamp at the bottom-right corner of each frame and explicitly instruct the model to use it as the temporal reference to identify the first accident frame. Given the resulting keyframe, the model then performs spatial localization by predicting bounding boxes for the involved subjects.

\textbf{(2) Traffic Accident Understanding (TAU)}: The model generates a factual description of the accident, infers plausible causes, and proposes avoidance suggestions.

\textbf{(3) Traffic Accident Responsibility Allocation (TARA)}: Finally, the model integrates traffic-law clauses retrieved via RAG to produce interpretable, evidence-grounded, and legally consistent responsibility attribution with self-verification.

Each step above is separately fine-tuned prior to integration, ensuring accuracy and stability throughout the chain and yielding reliable responsibility allocation in TARA.

\subsubsection{Retrieval-Augmented Generation (RAG)}
To enable responsibility reasoning grounded in legal principles, the AITP framework integrates an external traffic regulation knowledge base, which provides structured clauses from relevant traffic laws and regulations. During inference, the model retrieves and incorporates these legal texts to ensure that its responsibility judgments are both interpretable and compliant with official standards.
Given the factual and causal description $x$, we first encode it as a query vector:
\[
\mathbf{q} = f(x),
\]
and compute its similarity with each regulation embedding $\mathbf{z}_i$:
\[
s_i = \frac{\mathbf{q}^\top \mathbf{z}_i}{\tau},
\]
where $\tau$ is a temperature coefficient controlling the sharpness of similarity scores.

Then, we select the top-$K$ clauses with the highest similarity:
\[
C(x) = \operatorname{Concat}\left(d_i \mid i \in \operatorname*{TopK}_i(s_i)\right),
\]
where $\operatorname*{TopK}_i(s_i)$ denotes the indices of the $K$ most relevant clauses, and $\operatorname{Concat}(\cdot)$ represents the concatenation operation that merges the selected clauses into a unified context. 

Finally, the retrieved legal context $C(x)$ is injected into the model’s prompt, guiding the subsequent \textbf{responsibility allocation reasoning} to align with real-world traffic regulations.

\section{Experiments}
\label{sec:intro}
We conducted experiments on two aspects: multi-task and responsibility allocation. Since 1,000 videos were used for LoRA training in stage 4, 5,254 videos were used for Traffic Accident Responsibility Allocation (TARA), 946 videos for Traffic Accident Detection (TAD), and 2,308 videos for Traffic Accident Understanding (TAU) testing. Results show that our model achieves superior stability and accuracy in complex traffic scenarios.\\
\subsection{Experimental Setup}

{\bf Evaluation Metrics.}
For the responsibility allocation task (TARA), we report the accuracy of the model's predicted responsibility category across both accident and non-accident scenarios.

For textual outputs, including factual description, cause reasoning, and avoidance advice, we adopt BLEU~\cite{papineni2002bleu,post2018call}, ROUGE~\cite{lin2004rouge}, BERTScore~\cite{zhang2019bertscore}, and MoverScore~\cite{zhao2019moverscore}.
To further assess reasoning quality, we additionally employ GPTEval, following the LLM-as-a-judge paradigm \cite{zheng2023judging}, where we use Qwen3-30B as the judge model to score the generated answers in terms of consistency, correctness, and level of detail.

For time location, which predicts the first accident frame, we compute AP@3, AP@5, and AP@10, where a prediction is considered correct if it falls within $\pm k$ frames of the ground truth.
For spatial location, we report Average Precision under IoU thresholds (AP@30, AP@50, AP@70) together with mean IoU (mIoU).\\
{\bf Baselines.}
We compare our model with a set of strong general-purpose multimodal large language models. Specifically, we evaluate against Gemma-3n~\cite{team2025gemma}, Qwen3-VL~\cite{bai2025qwen3}, Kimi-VL-A3B\cite{team2025kimi}, and InternVL 3.5\cite{wang2025internvl3}, which represent the state-of-the-art vision–language reasoning capability across both image and video understanding tasks.

\subsection{Quantitative Comparison}
Table~\ref{tab:all_tasks_combined} demonstrates the superior performance of AITP on the TARA, TAD, and TAU tasks. The key findings are as follows: 

\begin{enumerate}
    \item \textbf{TARA.} AITP achieves very high responsibility-allocation accuracy on \emph{accident} videos, indicating the effectiveness of our method. Although its accuracy on \emph{non-accident} videos is relatively lower, this is mainly because AITP frequently identifies high-risk traffic scenarios and tends to infer that an accident has occurred. In contrast, other models achieve higher accuracy on non-accident data simply because they often fail to perceive accidents at all and consistently output ``no accident.''

    \item \textbf{TAU.} AITP shows a slight advantage, suggesting that other models already possess strong general video-description capabilities.

    \item \textbf{TAD.} AITP significantly outperforms all other models, demonstrating a far stronger spatiotemporal understanding of traffic accidents. This highlights AITP’s ability to accurately capture both the timing of the accident and the roles of all involved participants.
\end{enumerate}

\begin{table}[t]
\centering
\caption{Qualitative evaluation of the AITP framework.}
\begin{tabular}{c|p{0.55\linewidth}}
\hline
\textbf{Model} & \textbf{Traffic regulations recall} \\
\hline

Qwen3-VL &
A. Pedestrian takes main responsibility. Pedestrians should pay attention to their surroundings.
\\
\hline

AITP with RAG &
A. Pedestrian takes main responsibility. \textbf{Traffic regulations:}
According to Article 62, pedestrians shall use pedestrian crossings or crosswalks when crossing intersections or roads; when crossing intersections without traffic lights or pedestrian crossings, or when crossing roads in sections without crosswalks, they shall proceed only after confirming it is safe to do so...
\\
\hline
\end{tabular}
\label{tab:tara_overall}
\end{table}

\subsection{Qualitative Comparison}
As shown in Fig.~\ref{figure1}, Qwen3-VL fails to detect the occurrence of most traffic accidents in videos. 
In contrast, our AITP framework accurately identifies the accident type, the precise moment of occurrence, 
and all involved subjects, and further provides reliable responsibility allocation. 

As illustrated in Table~\ref{tab:tara_overall}, even when Qwen3-VL occasionally notices anomalies in a video, 
it often struggles to correctly identify the subjects involved and assign liability. 
In comparison, AITP delivers precise responsibility judgments and can automatically retrieve 
and cite the relevant traffic regulations.

\begin{table}[t]
\centering
\caption{Ablation results on progressive training and inference strategies, including the effects of DecaTARA fine-tuning, MCoT, and RAG.}
\setlength{\tabcolsep}{9pt}
\resizebox{\linewidth}{!}{
\begin{tabular}{l|ccc}
\toprule
\textbf{Method} & \textbf{Accident} & \textbf{Non-Accident} & \textbf{Average} \\
\midrule
\multicolumn{4}{l}{\textbf{Training Strategy}} \\
\midrule
Qwen3-VL (AITP\_stage0)          & 0.1895          & \textbf{0.9937} & 0.4956          \\
AITP\_stage1       & 0.3558          & 0.9380          & 0.5830          \\
AITP\_stage2       & 0.5206          & 0.8634          & 0.6568          \\
AITP\_mixtrained   & 0.6086          & 0.8019          & 0.6841          \\
AITP w/o CoT (AITP\_stage3)       & \textbf{0.7218} & 0.7771          & \textbf{0.7434}          \\
\midrule
\multicolumn{4}{l}{\textbf{Inference Strategy}} \\
\midrule
Qwen3-VL-CoT       & 0.2306          & \textbf{0.9912}          & 0.5274          \\
AITP\_implicit\_CoT & 0.7007          & 0.7800          & 0.7316          \\
AITP\_CoT              & 0.7218          & 0.8117          & 0.7569          \\
AITP\_CoT\_RAG      & \textbf{0.7272}          & 0.8171          & \textbf{0.7623}          \\
\bottomrule
\end{tabular}}
\label{tab:ablation}
\end{table}

\subsection{Ablation Study}
\textbf{Training strategy.} We first analyze the upper part of Table~\ref{tab:ablation}. For fair comparison, all models in this group are further equipped with Stage 4 LoRA fine-tuning. The results show that progressive training is more effective than directly optimizing all tasks together. Joint training is less effective because the supervision signals across tasks are heterogeneous: easier tasks may dominate optimization, while harder reasoning-intensive tasks are more difficult to learn directly. In contrast, progressive training improves the model’s accident understanding step by step through TAD, TAU, and TARA, moving from accident perception to deeper accident understanding and responsibility reasoning. This gain is not simply caused by overfitting to accident cues, since TARA requires not only recognizing whether an accident occurs, but also selecting the correct responsibility outcome from three candidate options. Meanwhile, the accuracy on non-accident videos decreases as training proceeds, mainly because the model becomes more sensitive to accident-related cues and tends to classify near-miss or other high-risk scenarios as actual accidents. By contrast, Qwen3-VL achieves high non-accident accuracy largely because it defaults to predicting ``no accident'' in most cases. In practice, this trade-off is acceptable, since missing a true accident is generally more problematic than over-predicting an accident in a challenging non-accident scenario.\\
\textbf{Inference strategy.} We then analyze the lower part of Table~\ref{tab:ablation}. For the base Qwen3-VL without DecaTARA fine-tuning, directly applying CoT already improves performance, showing that structured reasoning is helpful even without task-specific training. After progressive fine-tuning, explicit multi-step CoT further outperforms direct inference, indicating that step-by-step reasoning helps the model verify accident existence before making responsibility judgments. By contrast, implicit CoT performs worse because all reasoning steps are compressed into a single interaction. Without explicit accident-frame extraction and spatial localization, and with an overly long prompt, the reasoning process becomes less clear and the final decision becomes less stable. Incorporating RAG brings further gains, showing that retrieved traffic regulations reinforce the model’s judgment on key responsibility criteria and improve final responsibility allocation.

\section{Conclusion}
We present AITP, a multimodal large language model for traffic accident responsibility allocation, together with DecaTARA, a ten-task benchmark covering accident perception, temporal-spatial understanding, causal reasoning, and regulation-grounded decision-making. Experiments show that AITP consistently outperforms general MLLMs across all tasks, achieving strong accident understanding and reliable responsibility allocation. Our results also highlight the value of progressive training, multimodal chain-of-thought reasoning, and legal knowledge retrieval. We hope AITP and DecaTARA can support future research on safety-critical multimodal reasoning and law-aligned AI systems.
\section*{Acknowledgment}
This work was supported by the National Natural Science Foundation of China under Grant 52402504, and by the GIFT Future Scholars Program.
{
    \small
    \bibliographystyle{ieeenat_fullname}
    \bibliography{main}
}


\end{document}